\documentclass[conference]{IEEEtran}
\IEEEoverridecommandlockouts
\usepackage[left=0.75in,right=0.75in,top=0.75in, bottom=0.75in]{geometry}
\usepackage[T1]{fontenc}
\usepackage{cite}
\usepackage{amsmath,amssymb,amsfonts}
\usepackage{algorithmic}
\usepackage{graphicx}
\usepackage{textcomp}
\usepackage{xcolor}
\usepackage{booktabs}
\usepackage{subcaption}
\usepackage{url}

\def\BibTeX{{\rm B\kern-.05em{\sc i\kern-.025em b}\kern-.08em
    T\kern-.1667em\lower.7ex\hbox{E}\kern-.125emX}}

\begin{document}
\title{Terrain Classification for the Spot Quadrupedal Mobile Robot Using Only Proprioceptive Sensing\\\thanks{This project was funded in part by the Natural Sciences \& Engineering Research Council of Canada (NSERC), Defence R\&D Canada (DRDC), and General Dynamics Land Systems (GDLS) under project DNDPJ 533392-18 as well as USRA.}
}

\author{\IEEEauthorblockN{Sophie Villemure}
\IEEEauthorblockA{\textit{Ingenuity Labs Research Institute} \\
\textit{Queen's University}\\
Kingston, Canada \\
20svv2@queensu.ca}
\and
\IEEEauthorblockN{Jefferson Silveira}
\IEEEauthorblockA{\textit{Ingenuity Labs Research Institute} \\
\textit{Queen's University}\\
Kingston, Canada \\
jefferson.silveira@queensu.ca}
\and
\IEEEauthorblockN{Joshua A.\ Marshall}
\IEEEauthorblockA{\textit{Ingenuity Labs Research Institute} \\
\textit{Queen's University}\\
Kingston, Canada \\
0000-0002-7736-7981}
}

\maketitle

\begin{abstract}
Quadrupedal mobile robots can traverse a wider range of terrain types than their wheeled counterparts but do not perform the same on all terrain types. These robots are prone to undesirable behaviours like sinking and slipping on challenging terrains. To combat this issue, we propose a terrain classifier that provides information on terrain type that can be used in robotic systems to create a traversability map to plan safer paths for the robot to navigate. The work presented here is a terrain classifier developed for a Boston Dynamics Spot robot. Spot provides over 100 measured proprioceptive signals describing the motions of the robot and its four legs (e.g., foot penetration, forces, joint angles, etc.). The developed terrain classifier combines dimensionality reduction techniques to extract relevant information from the signals and then applies a classification technique to differentiate terrain based on traversability. In representative field testing, the resulting terrain classifier was able to identify three different terrain types with an accuracy of approximately 97\%

\end{abstract}


\section{Introduction}
Quadrupedal mobile robots can traverse a wider range of terrain types than their wheeled counterparts. However, quadrupedal robots are prone to undesirable behaviours like sinking and slipping on challenging terrain. Non-geometric hazards tied to terrain types should play a role in path and gait planning to achieve optimal behaviour. Terrain classification can identify hazards and inform both path and gait planning.

    This paper presents an approach to terrain classification that relies only on proprioceptive sensing. Such sensors are those that measure the state of the robot itself. Examples include position, speed, and joint angles. In contrast, exteroceptive sensors directly measure the robot's surrounding environment (e.g., camera images or LiDAR point clouds). One advantage of our reliance on only proprioceptive sensing is that the proposed approach is more robust to changes in lighting factors. The proposed approach also utilizes only pre-existing proprioceptive sensors on a Boston Dynamics Spot robot. 

\subsection{Related Work}
There are many approaches to terrain classification presented in the literature. Cameras are the most common sensors used for terrain classification since vision data can provide information at a further range than other sensor types. Vision-based systems use data from stereo-vision or monocular cameras to extract information from the environment \cite{gao2014survey}. Cameras can be combined with LiDAR data and computer vision methods to classify terrain. Shirkhodaie et al.\ \cite{shirkhodaie2005soft} explore several vision-based terrain classifiers including a Heuristic Rule-Based Classifier (HRBC), a Neural Network-based Terrain Classifier (NNTC) and a Fuzzy Logic-based Classifier. They achieved terrain classification accuracies in the 90 \% range for the latter two of the three classifiers. Vision-based methods have the advantage of being able to classify terrain from a distance, but they are limited by the quality of the camera, the visibility and the lighting conditions. Machine learning classification methods have proven to create classifiers with high accuracy but have relatively high computational complexity compared to classical approaches.

Another approach to terrain classification is tactile-based terrain classification. Ding et al.\ \cite{ding2022pressing} used a haptic-based approach to classify terrain for a legged robot. Their haptic approach uses two tactile motions to extract features of the walking surface (e.g., friction, softness). The features are processed and analyzed with a few different classical classifiers including K-Nearest Neighbours (KNN), Support Vector Machine (SVM), Gaussian Discriminant Analysis (GDA) and Logistic Regression (LR). This tactile approach can differentiate between 12 different terrain types with accuracy in the 80-95 \% range. Tactile approaches to terrain classification can not classify terrain from a distance but are not affected by lighting conditions and have a lower computational complexity than vision-based methods. The haptic approach does require the addition of extra sensors to the robot and requires the robot to use a specific gait. In addition to these disadvantages, since legged robots are not continuously in contact with the ground many tactile approaches require longer walking samples per classification to extract meaningful data \cite{hoepflinger2010haptic}. 

Iagnemma et al.\ \cite{Iagnemma2005Vibration} was the first to use a vibration-based approach to classify terrain for a wheeled robot, this approach uses features extracted from an IMU attached to the axle of the robot's wheel to classify the terrain. Their work aimed to classify two terrain types, but the concept has been developed by several other researchers. Weis et al.\ \cite{weiss2006vibration} use an SVM classifier on features extracted from an IMU on a cart and can classify up to seven classes while maintaining misclassification rates below 7 \%. Vibration-based approaches are not affected by lighting conditions and do not necessarily require a dedicated sensor.

Zhao et al.\ \cite{zhao2017new} proposed a framework for terrain classification that combines vibration-based and acoustic-based classification strategies. Their work demonstrates that proprioceptive classification strategies are a valuable tool to enhance other approaches to terrain classification. 

Except for the tactile-based approach discussed, most approaches to terrain classification in literature are developed for wheeled mobile robots. For some of these methods, such as vision-based approaches, the performance of the classifier is independent of the structure of the robot. Therefore we expect these classifiers to perform similarly on quadrupedal robots. A far more interesting problem is trying to create a classifier that leverages the unique aspects of quadrupedal robots. 

\subsection{About this Paper}
This paper proposes a method that takes advantage of the unique proprioceptive data that is available to legged robots and adapts feature extraction techniques from vibration-based classification methods. The feature selection, reduction and classification techniques used in this approach are reminiscent of those used in tactile-based techniques. Our proposed algorithm involves two phases; an offline training phase and an online classification phase. The offline phase involves data processing, feature selection, dimensional reduction and training. The online phase uses the features, dimensional reduction transform and transformed dataset from the offline phase to classify terrain in real-time. The algorithm was able to classify three terrain types (concrete, grass, and rocks) in experiments containing single and mixed terrain types, with a classification accuracy up to 97 \%. The source code, along with a video demonstrating the terrain classification, is accessible at \url{https://github.com/offroad-robotics/terrain_classifier}.

This paper is organized as follows. Section II discusses the proposed approach to terrain classification, Section III presents our methodology for tuning and testing the classifier, Section IV presents the offline and online test results, and Section V concludes the paper and discusses future work.
\section{Terrain Classification Algorithm}
This section presents an overview of the framework used to classify the terrain traversed by a quadrupedal robot. The proposed method includes the following steps:
\begin{enumerate}
    \item \label{item:collect} Collect and process raw data from the robot's proprioceptive sensors;
    \item \label{item:feature} Conduct Feature Selection using a min-Redundancy, Max-Relevance (mRMR) algorithm \cite{8964172};
    \item \label{item:reduce} Reduce the dimensionality of the dataset using Principal Component Analysis (PCA); and,
    \item \label{item:classify} Classify the terrain with a K nearest neighbours (KNN) classifier and the transformed dataset.
\end{enumerate}

The biggest challenge with adapting the work done on wheeled robots for quadrupedal robots is that quadrupedal robot systems, like Spot, have more degrees of freedom. This results in more signals that can be used for classification and greatly increases the complexity of the classification problem. The signals are difficult to interpret compared to, for example, vibration signals used on wheeled robots. On the other hand, quadrupedal robots that contain multiple internal sensors, such as the Boston Dynamics Spot robot, provide more information that can be used to potentially improve the accuracy of a terrain classifier algorithm based solely on proprioceptive data.

\begin{figure}
    \centering
    \includegraphics[width=0.91\columnwidth]{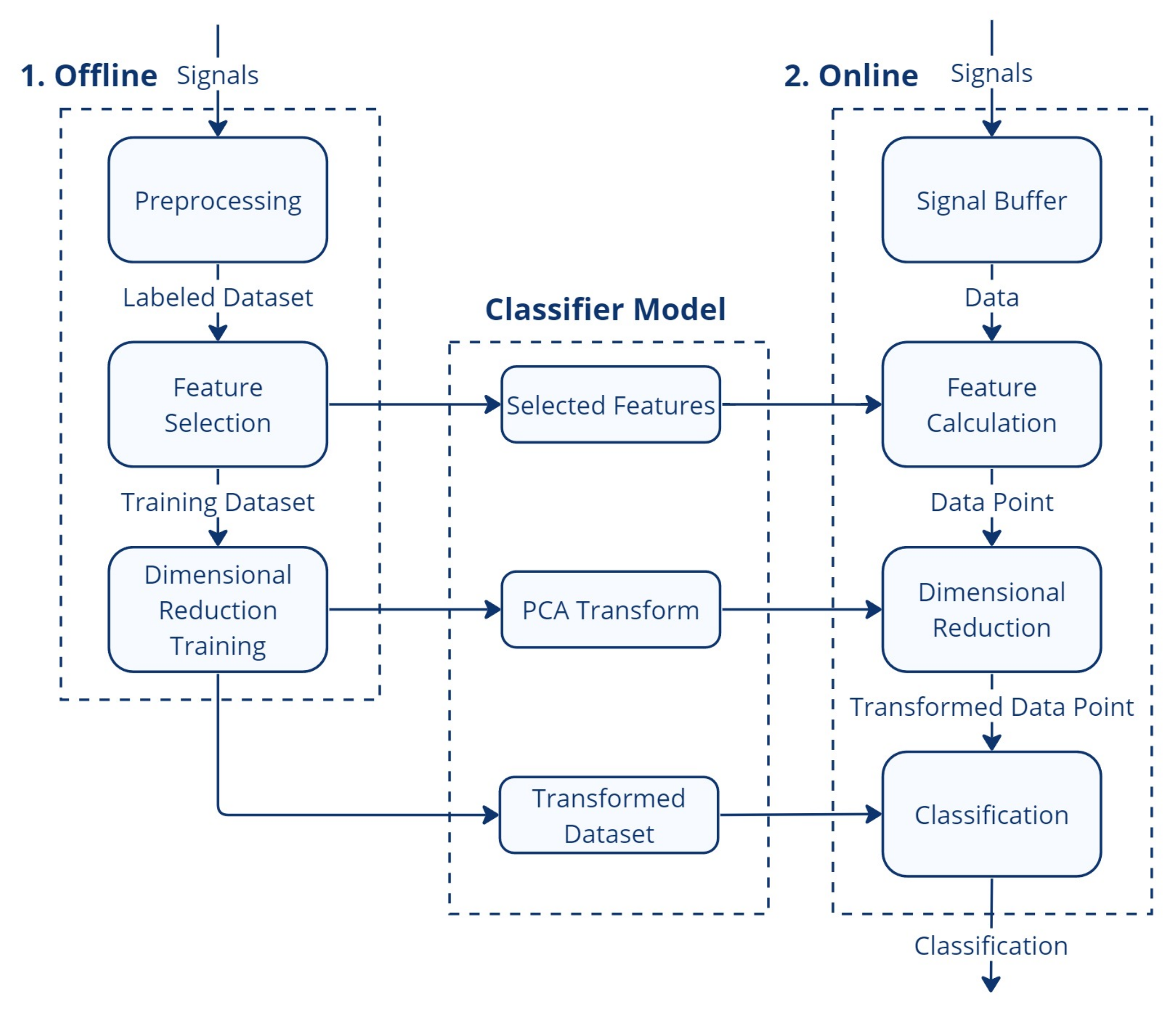}
    \caption{Diagram of the offline and online phases of the proposed method.}
    \label{fig:method}
\end{figure}

Our proposed approach aims to reduce the problem's dimensionality as much as possible while maintaining necessary and relevant information. This strategy strives to automatically filter irrelevant data, simplify the algorithm, and reduce the amount of required training data and resources without compromising the accuracy of the algorithm.  Fig.~\ref{fig:method} shows a block diagram of the proposed method. The offline phase of the method includes Steps \ref{item:collect} to \ref{item:reduce} and the online phase uses information derived from the offline phase to complete Step~\ref{item:classify}. 

\subsection{Preprocessing}
\begin{figure}[t]
    \centering
    \includegraphics[width = 0.7\columnwidth]{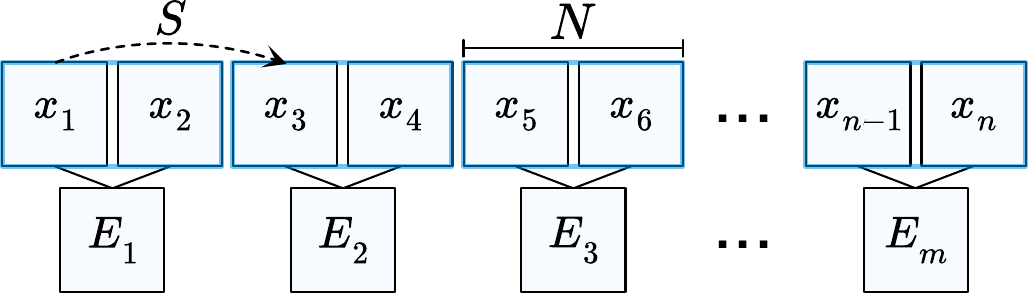}
    \caption{Example of energy computation for $S = 2$ and $N = 2$. }
    \label{fig:energy_calculation}
\end{figure}

We employed Boston Dynamics' Robot Operating System (ROS) interface, which provides over 144 different time series signals related to the position and speed of the robot, each of its 12 joints and each of its four feet. In the preprocessing stage, to train the classifier, the signals are labelled based on terrain type. Then, the ``energy'' of the signal is calculated using an approach similar to a moving average filter with window size $N$ and stride $S$, as shown in Fig. \ref{fig:energy_calculation}. We match the stride value to the window size ($S=N$) to down-sample the signal for training. $N$ is used as a hyperparameter to tune the classifier.

The signal energy $E_m$ is calculated by
\begin{equation}
    E_m = \sum_{k=1}^{N} (x_{(m-1)S + k} - \bar{x}_m)^2,
    \label{eq:signal_energy}
\end{equation}
where $\bar{x}_m$ is the moving average calculated as
\begin{equation}
    \bar{x}_m = \frac{1}{N}\sum_{k=1}^{N} [x_{(m-1)S + k}].
    \label{eq:signal_mean}
\end{equation}

We use the signal energy as an indicator of the "traversability" of a terrain with the hypothesis that a less traversable terrain will cause more vibrations, which increases the signal energy. Visual inspection of experimental data revealed that signal energy was indeed an indicator of terrain type for many signals. One exception to this was the ground penetration mean and standard deviation signals from the foot state information. For these signals, the moving average value \eqref{eq:signal_mean} was used as a feature since it was more indicative of terrain type.

\subsection{Feature Selection}
Selecting the features to use in the classifier is an important step in the classification process. However, selecting features manually proved to be a laborious and ineffective process. To streamline the selection of the most relevant features, we used an mRMR algorithm, which assigns a score to each feature representing the importance of that feature in distinguishing between different categories. More precisely, the proposed method uses an F-test Correlation Quotient (FCQ) mRMR algorithm. The scores assigned by the FCQ mRMR algorithm are the ratio of the redundancy of a feature (represented by the F-Statistic) to the relevancy of a feature (represented by the Pearson correlation). The FCQ score of a feature is
\begin{equation}
    FCQ = \frac{F(Y, X_i)}{\frac{1}{|S|} \sum_{X_s \epsilon S} {\rho(X_s, X_i)}}, 
    \label{mRMR score}
\end{equation}
where $F(Y, X_i)$ is the F-Statistic of feature $X_i$ and $\rho(X_s, X_i)$ is the Pearson correlation of feature \cite{8964172}.

The feature selection stage takes in the labelled features extracted during the preprocessing step, runs the FCQ mRMR and outputs the FCQ score associated with each feature. The features with the highest scores are then passed to the dimensional reduction step. The number of features selected is a tunable parameter of the classifier.  

\subsection{Dimensional Reduction}
Once the most relevant features have been selected, the next step is to reduce the dimensionality of the dataset, which allows us to reduce the computational complexity of the algorithm. We employ principal component analysis (PCA), which is a linear transformation that re-expresses the data according to a new set of eigenvalues and their corresponding eigenvectors named the principal components (PCs). The PCs are arranged such that the first PC retains more information than the second and so on. The percentage of the total information retained in a PC is called the explained variance. Since the last few PCs often retain very little information, we can choose to ignore them. Doing so can significantly decrease the dimensionality of a problem while retaining most of the explained variance. In this classifier, we determined the number of PCs to keep based on a given cumulative explained variance threshold. The cumulative explained variance threshold is a tunable parameter. The output of a PCA transform naturally forms clusters for each distinct terrain type since it is trained on labelled data. PCA requires a training dataset and in return, the algorithm provides a transformation matrix that can be used to transform data in real time. This classifier uses the PCA implementation developed by \cite{scikit-learn}.

\subsection{Classification}
The $k$-nearest neighbours (KNN) classifier is used as the last step to classify the terrain. This classifier was chosen for its simplicity, and it works by finding the $k$-nearest training points to the test point and then classifying the test point based on the majority class of the $k$-nearest training points. The KNN classifier implementation uses the Euclidean distance. The number $k$ is a tunable parameter of the system. The accuracy of the KNN classifier reduces as the number of classification categories increases.  
\section{Methodology}
To evaluate the proposed algorithm, we ran tests offline on collected data, as well as tests in real time with a Boston Dynamics Spot robot. This section outlines the methodology used to collect data, tune the classifier, and test its accuracy on data that was not used to train the classifier model, shown in Fig.~\ref{fig:method}. For this discussion, the accuracy of the classifier is defined as the percentage of samples that were correctly classified. The goal of our testing is to evaluate the accuracy of the proposed classification method, the real-time performance of the proposed method, and the sensitivity of the proposed approach to tuning, velocity and other environmental factors. 

\begin{figure*}[t]
    \centering
    \begin{subfigure}{0.2\textwidth}
        \centering
        \includegraphics[width = 1\linewidth]{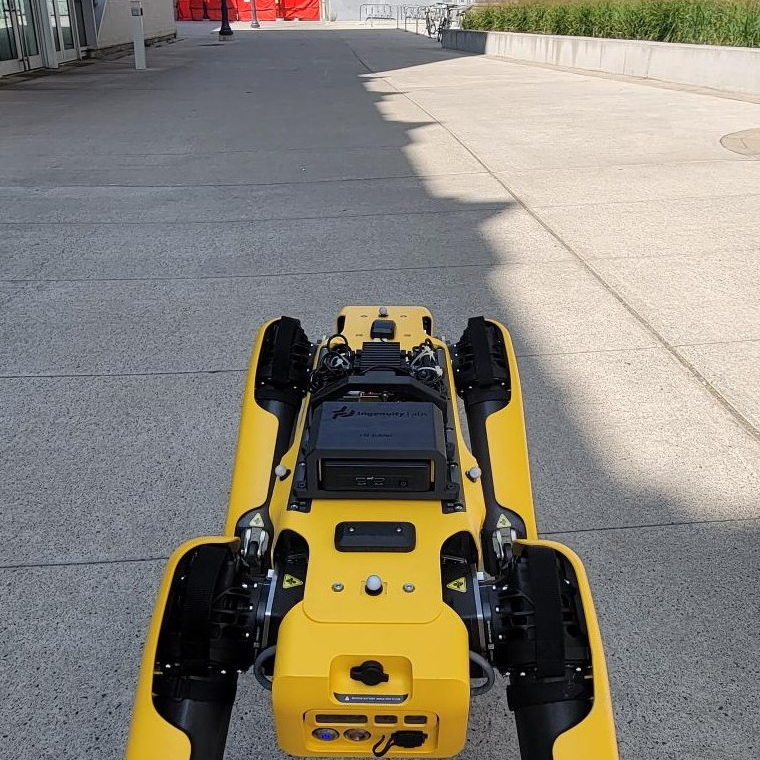}
        \caption{}
        \label{fig:concrete}
    \end{subfigure}
    \begin{subfigure}{0.2\textwidth}
        \centering
        \includegraphics[width = 1\linewidth]{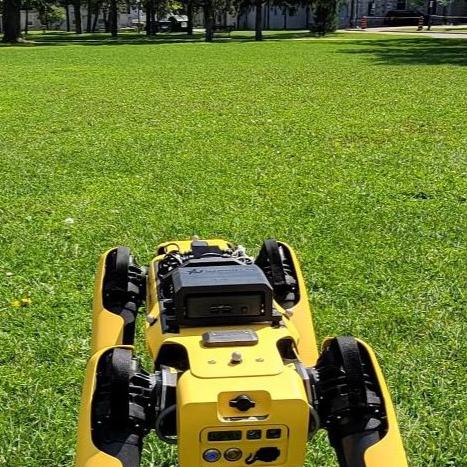}
        \caption{}
        \label{fig:grass}
    \end{subfigure}
    \begin{subfigure}{0.2\textwidth}
        \centering
        \includegraphics[width = 1\linewidth]{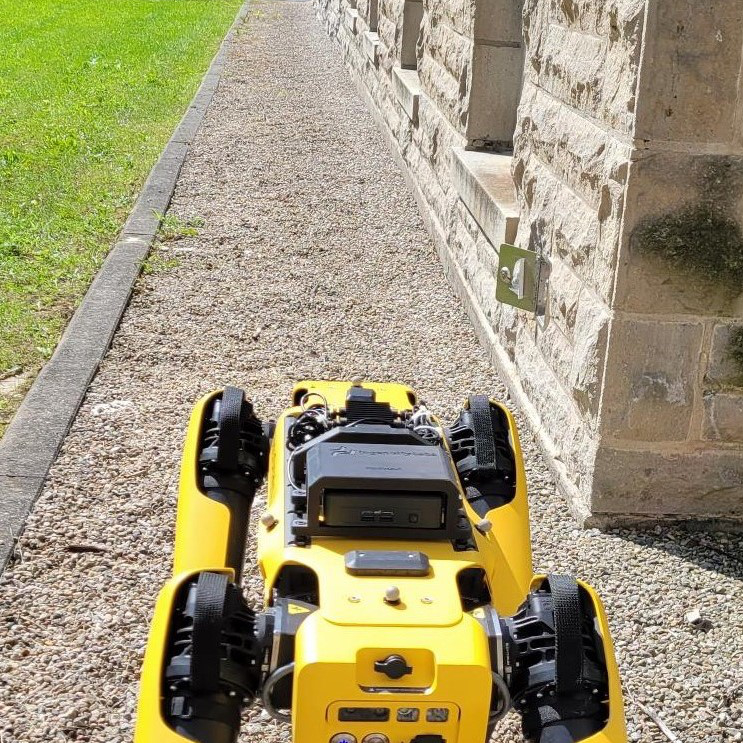}
        \caption{}
        \label{fig:rocks}
    \end{subfigure}
    \begin{subfigure}{0.2\textwidth}
        \centering
        \includegraphics[width = 1\linewidth]{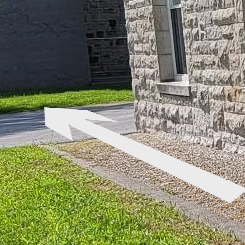}
        \caption{}
        \label{fig:mixed_terrain_arrow}
    \end{subfigure}
    \caption{Classified terrains: (a) concrete (b) grass, and (c) rocks. In (d) we show the mixed terrain used for evaluation. The white arrow indicates the direction taken by the robot.}
    \label{fig:training_terrain}
\end{figure*}

\subsection{Data Collection}
Data was collected using a Boston Dynamics Spot robot and the Spot ROS interface. The Spot ROS interface provides access to the IMU data, joint state information and foot state information at a rate of 16.7~Hz. The robot was taken to three sites on the Queen's University Campus, Figs.~\ref{fig:concrete} to \ref{fig:rocks} show the three sites used for training data collection.

The robot walked over each site at a constant speed. 
The validation data was a randomly selected subset of the training data set aside and not used in the training process. The validation data was used for hyperparameter tuning. 

All data was collected on flat terrains, around 30 seconds of data was collected for each terrain type for training and validation, and 1 minute of data was collected for each terrain type for testing. The data for offline testing was taken at the same sites as the training data but on a different day. For testing, the validation data was included in the training process. 

In addition to the testing dataset, a mixed terrain dataset was collected. Fig.~\ref{fig:mixed_terrain_arrow} shows the site used for testing. The mixed terrain dataset was collected by walking the robot forward at a speed of 1 m/s across the three tested terrain types in succession. This site was not the same site on which the classifier was trained.

\subsection{Hyperparameter Tuning}
The classifier has four tunable parameters: 1) the window size $N$ used in preprocessing; 2) the number of features retained by the feature selection stage; 3) the explained variance threshold for the dimensional reduction stage; and, 4) the number of neighbours for the KNN classifier. The parameters were tuned to optimize accuracy on the validation dataset. The values selected for the final classifier were a window size of 9 frames (0.5~s), 22 mRMR features, an explained variance threshold of 65~\% and 5 neighbours. These values resulted in an accuracy of around 95~\% on the validation set.

\subsection{Offline Testing}
The offline tests closely resembled the validation tests used for tuning. The data was collected, extracted and processed in the same way as the training and validation data. The classifier was only trained at a speed of 1~m/s but the test dataset included data at different speeds to test the effect of walking speed on classification. The Offline testing used the same classifier model as the online testing. 

\subsection{Online Testing}
Since the proposed method uses a simple classifier with a small training set, the classifier can run in real time onboard the robot. Instead of splitting the data into windows, the classifier stores the last few data points (equivalent to the window size) in a circular buffer and uses the points to classify the terrain. This is equivalent to calculating the mean and energy in \eqref{eq:signal_mean} and \eqref{eq:signal_energy} with $S = 1$. The classifier only saves the signals that are used for classification, which reduces memory usage. Features are then extracted from the signals and transformed with the PCA transform established during training. The transformed features are then classified with the KNN classifier.

Real-time classification was tested by having an operator walk the robot on various terrains, in contrast with offline tests that used the robot's autonomy system to command the speed. The Spot Controller comes with three speed settings that affect the maximum speed of the robot. Slow, normal, and fast settings limit the robot's speed to 0.5~m/s, 1~m/s and 1.5~m/s, respectively. The classifier was tested on all three terrain types at all three speed settings. Since the operator used a handheld controller, the walking speed was not constant and the robot was not always travelling in a straight line. 
\section{Experimental Results}

\subsection{Offline Tests}
\begin{figure}
    \begin{subfigure}[t]{0.48\columnwidth}
        \includegraphics[width=\linewidth]{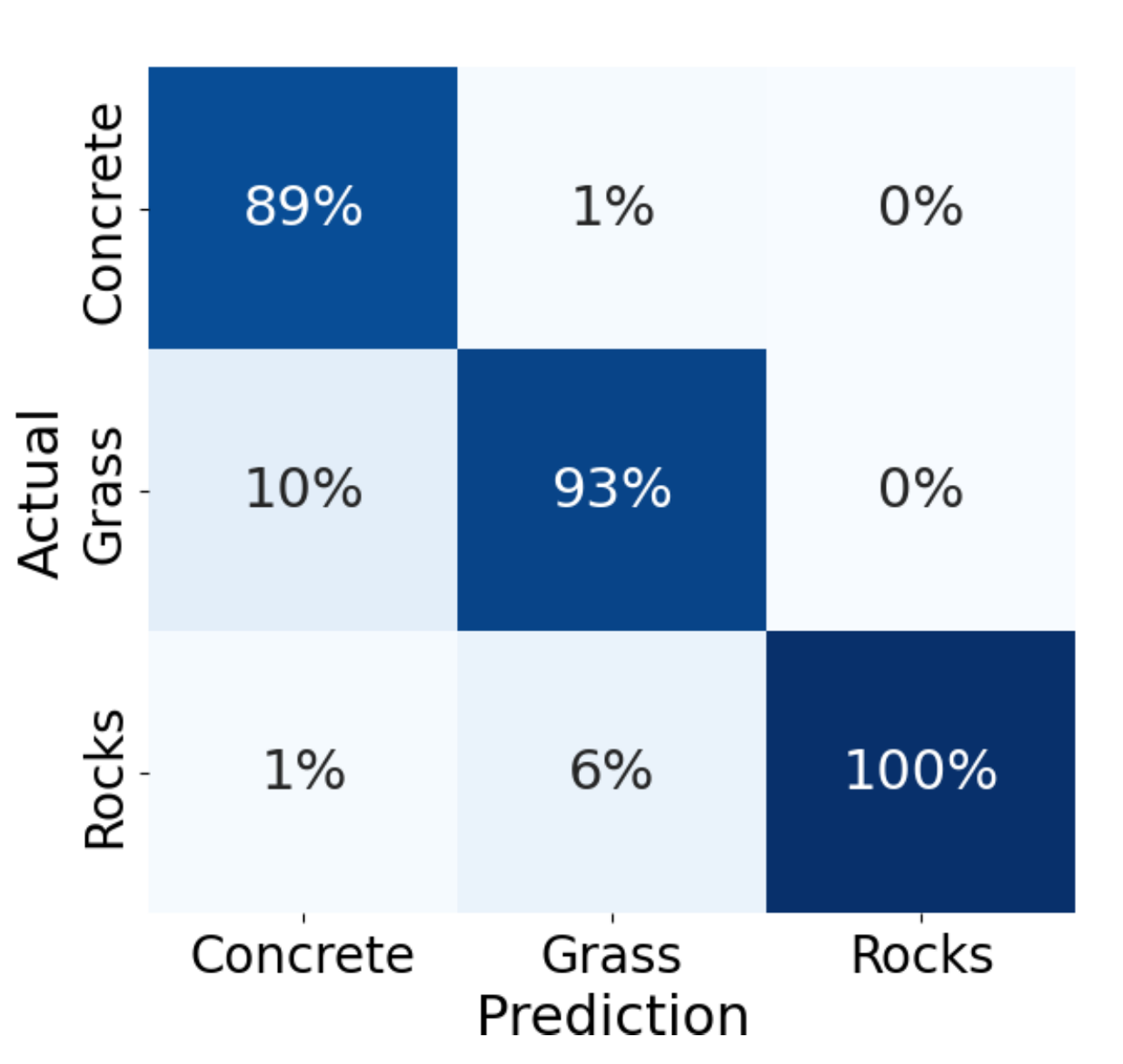}
        \caption{Offline at 0.5 m/s}
        \label{fig:confusion matrix 0.5}
    \end{subfigure}
    \hfill
    \begin{subfigure}[t]{0.48\columnwidth}
        \includegraphics[width=\linewidth]{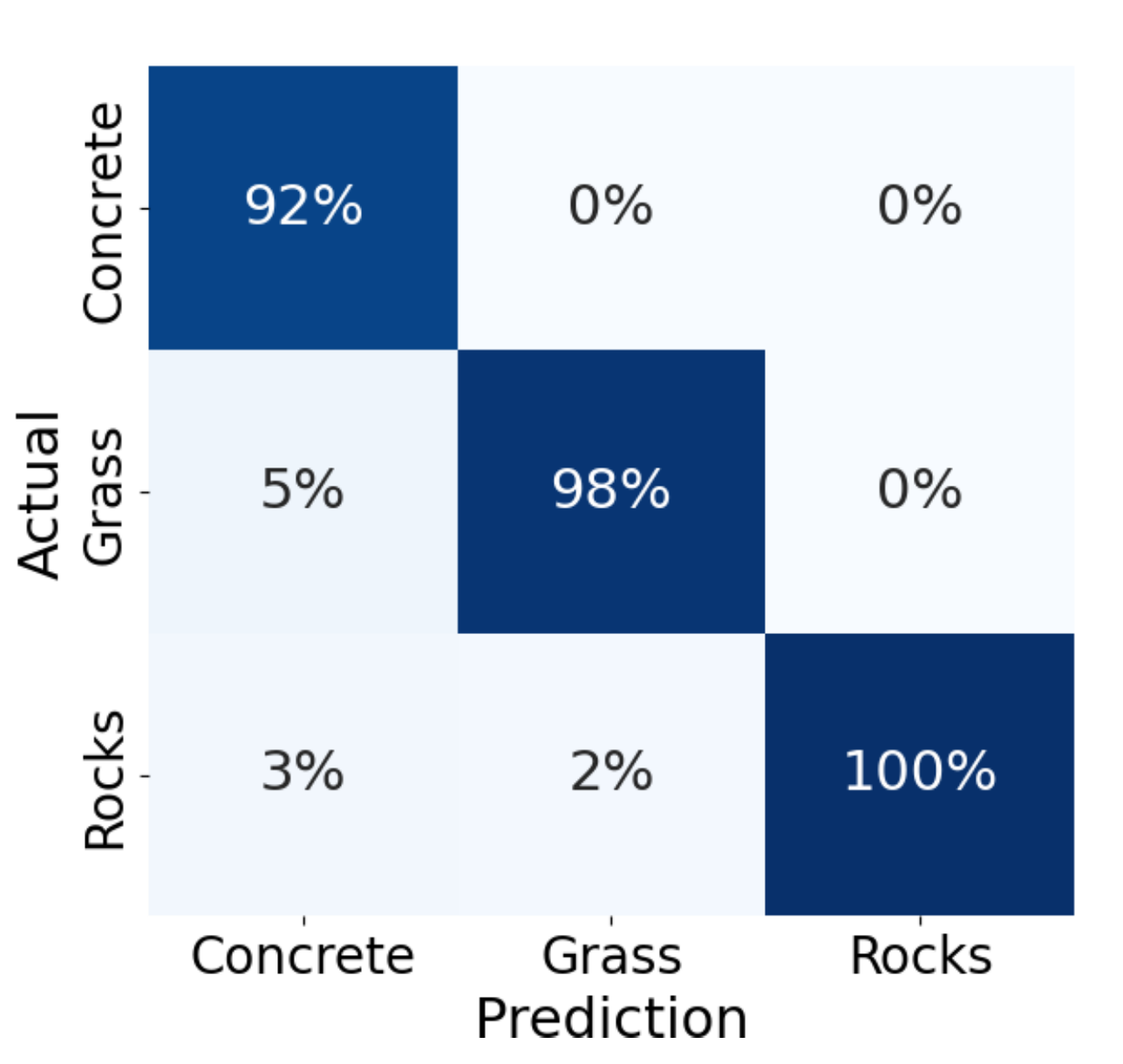}
        \caption{Offline at 1 m/s (training speed)}
        \label{fig:confusion matrix}
    \end{subfigure}
    \vspace{\baselineskip}
    \begin{subfigure}[t]{0.48\columnwidth}
        \includegraphics[width=\linewidth]{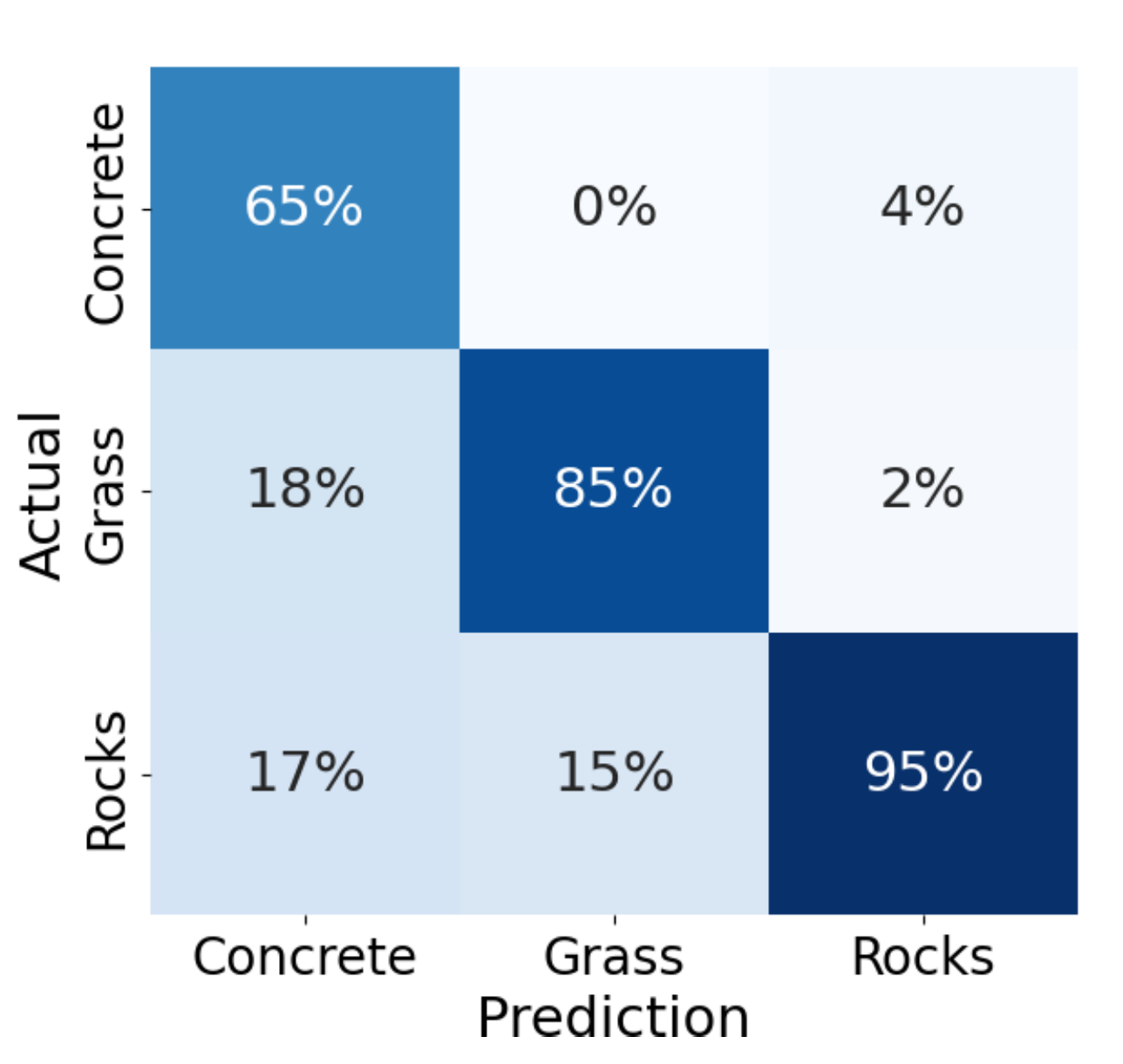}
        \caption{Online at a slow speed}
        \label{fig:online confusion matrix 0.5}
    \end{subfigure}
    \hfill
    \begin{subfigure}[t]{0.48\columnwidth}
        \includegraphics[width=\linewidth]{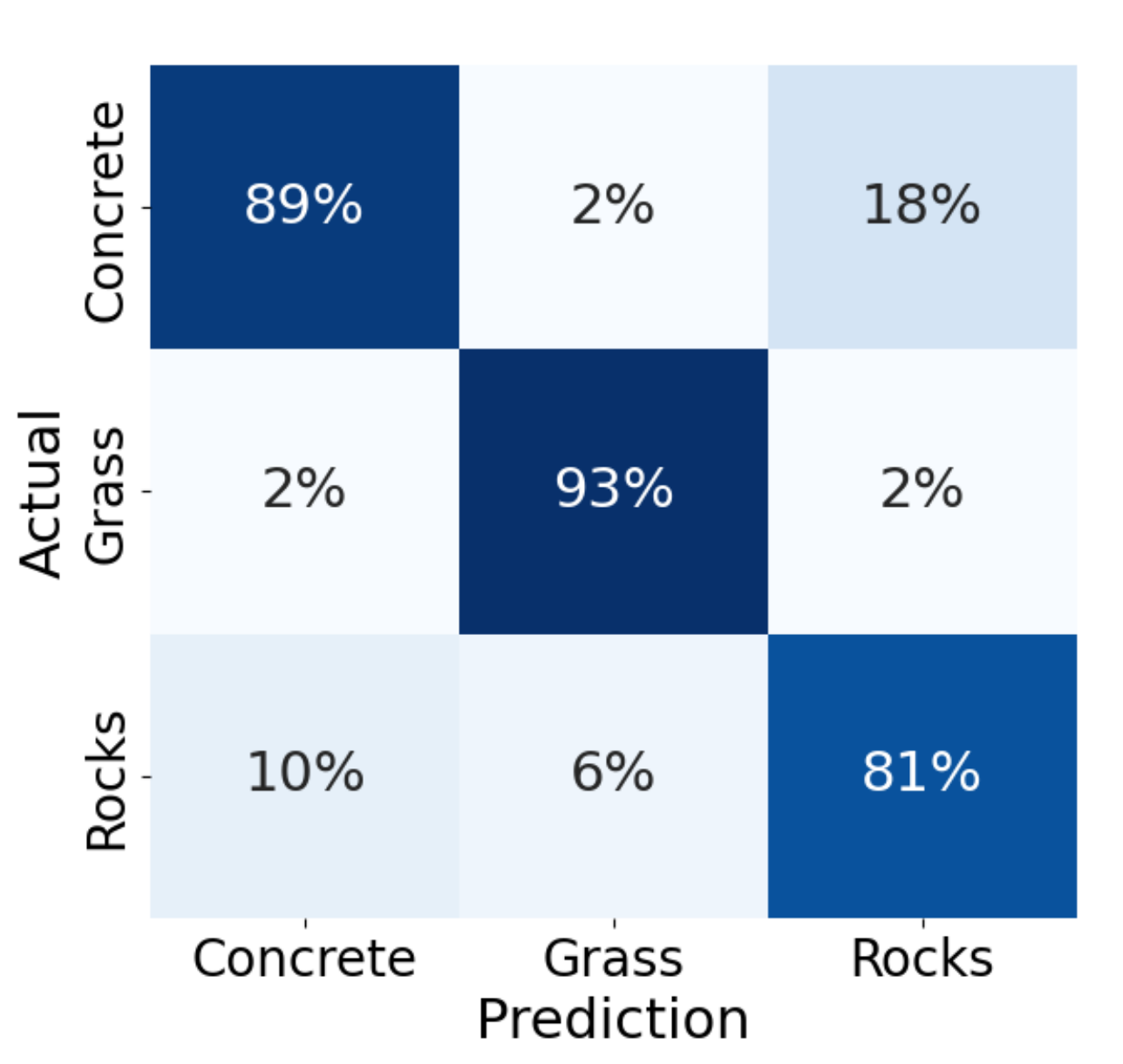}
        \caption{Online at a normal speed (close to the training speed)}
        \label{fig:online confusion matrix}
    \end{subfigure}
    \caption{Confusion matrices showing the classifier performance.}
    \label{fig:performance}
\end{figure}

Fig.~\ref{fig:confusion matrix} shows the confusion matrix for the classifier on the test dataset at 1~m/s. The confusion matrix indicates that concrete was classified as grass 5\% of the time. The overall accuracy of the classifier at this speed was 97~\%. Fig.~\ref{fig:confusion matrix 0.5} shows the confusion matrix for the classifier on the test dataset at 0.5~m/s. The accuracy of the classifier on the test dataset at 0.5~m/s was 94~\%. The slight drop in accuracy demonstrates that the speed of travel had a small effect on the accuracy of the classifier.

\begin{figure}
    \centering
    \begin{subfigure}[t]{0.49\columnwidth}
        \centering
        \includegraphics[width = 1\linewidth]{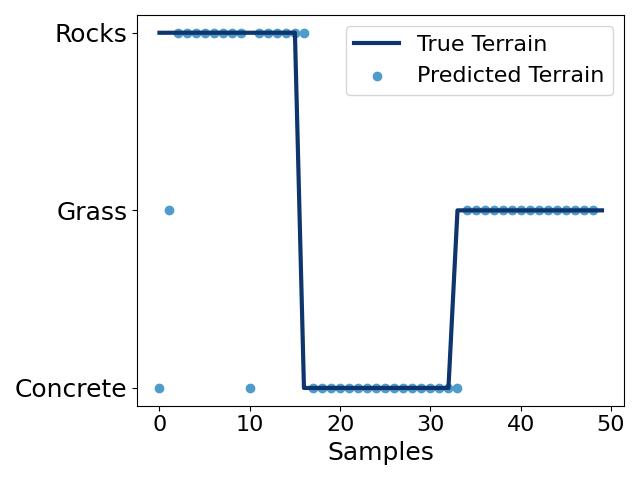}
        \caption{Offline performance with the terrain pictured in Fig. \ref{fig:mixed_terrain_arrow}}
        \label{fig:mixed terrain results}
    \end{subfigure}
    \begin{subfigure}[t]{0.49\columnwidth}
        \centering
        \includegraphics[width = 1\linewidth]{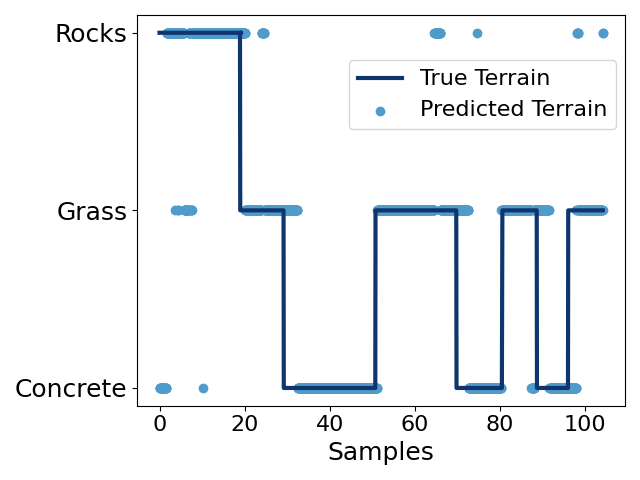}
        \caption{Online performance}
        \label{fig:online mixed terrain results}
    \end{subfigure}
    \caption{Classification performance with mixed terrains.}
    \label{fig:classification_performance}
\end{figure}

The classifier was also tested on a mixed terrain dataset. Fig.~\ref{fig:mixed terrain results} shows the results of one of the trials on the mixed terrain dataset. The average accuracy of the classifier on the mixed terrain dataset was on average 86~\%. Most of the misclassifications occurred when the robot was walking on rocks. There was a small but noticeable lag between when the robot transitioned from one terrain type to another and when the classifier identified the change in terrain. 

\subsection{Online Tests}
Fig.~\ref{fig:online confusion matrix 0.5} and Fig.~\ref{fig:online confusion matrix} show the confusion matrices of the online tests at slow and normal speeds. The overall accuracy of the classifier at each speed was 75~\% and 87~\%respectively. The classifier was tested at fast speeds but performed poorly. The classifier unsurprisingly performed the best at the training speed. At the slow speed, the classifier tended to under-classify the data. This is likely because the speed of travel directly affected the estimated energy, which is the primary feature used in the classifier. It might be possible to obtain better results if the window size $N$ is selected based on the speed of travel.

The most significant difference between the online and offline tests is that the robot was not travelling at a constant speed, nor was it travelling in a straight line. The performance of the online tests shows that the classifier performs well as the robot turns or experiences minor variations in speed.

Fig.~\ref{fig:online mixed terrain results} shows the results of the online mixed terrain test. Like in the offline mixed terrain test in Fig.~\ref{fig:mixed terrain results}, there was a noticeable lag between when the robot changed terrain and when the classifier identified the change. The lag was more noticeable in the online test than in the offline test, but this was likely due to the online test having more classifications per second than the offline test. The accuracy of the classifier during this test was 82~\%, and most of the misclassifications were at the transitions between terrain types.
\section{Conclusion}
This paper presents a new approach to terrain classification for the Spot quadrupedal mobile robot using only its factory-installed proprioceptive sensors. From the signals, we extract signal energy and average value features and use an mRMR algorithm to select the most relevant features. A PCA algorithm is used to reduce the number of dimensions further while maintaining 65\% of the initial information. In our outdoor experiments, this process reduced the number of features from over 100 to three, which allowed us to use a simple KNN classifier to classify terrains. The classifier achieved a high degree of accuracy while maintaining a low computational complexity. The low computational load of the final classifier allowed it to run in real-time onboard the robot. Despite the small dataset used to train the classifier, the classifier performed well on a mixed terrain dataset including terrain it was not trained on.

The chosen features presented some limitations, since signal energy is related to the speed of the robot, the classifier frequently misclassified the terrain when the robot was walking at a different speed than the training set. Some other features may be less susceptible to this effect. Future work on this problem will explore other feature types and classification methods that may be more robust to changes in speed , future work will also explore how the classifier responds to hills and uneven terrain.

\bibliographystyle{IEEEtran}
\bibliography{references}
\end{document}